\documentclass[conference]{IEEEtran}
\IEEEoverridecommandlockouts
\usepackage{cite}
\usepackage{amsmath,amssymb,amsfonts}
\usepackage{algorithmic}
\usepackage{graphicx}
\usepackage{textcomp}
\usepackage{xcolor}
\usepackage{hyperref}
\usepackage{booktabs}
\usepackage{gensymb}
\usepackage{lipsum}
\usepackage{subcaption}
\def\BibTeX{{\rm B\kern-.05em{\sc i\kern-.025em b}\kern-.08em
    T\kern-.1667em\lower.7ex\hbox{E}\kern-.125emX}}
\begin{document}

\title{Score-Based Multibeam Point Cloud Denoising\\
% Acknowledgement added at the end of the paper
% \thanks{*This work was supported by Swedish Maritime Robotics Center (SMaRC).}
}

\author{Li Ling$^{1}$, Yiping Xie$^{1}$, Nils Bore$^{2}$, John Folkesson$^{1}$
\thanks{$^{1}$Division of Robotics, Perception and Learning (RPL), KTH Royal Institute of Technology, Stockholm, Sweden
        {\tt\small \{liling,yipingx,johnf\}@kth.se}}%
\thanks{$^{2}$Ocean Infinity, Sven Källfelts Gata 11, SE-426 71 Västra Frölunda, Sweden
        {\tt\small nils.bore@oceaninfinity.com}}%
}

\maketitle

\begin{abstract}
Multibeam echo-sounder (MBES) is the de-facto sensor for bathymetry mapping. In recent years, cheaper MBES sensors and global mapping initiatives have led to exponential growth of available data. However, raw MBES data contains $1-25\%$ of noise that requires semi-automatic filtering using tools such as Combined Uncertainty and Bathymetric Estimator (CUBE). In this work, we draw inspirations from the 3D point cloud community and adapted a score-based point cloud denoising network for MBES outlier detection and denoising. We trained and evaluated this network on real MBES survey data. The proposed method was found to outperform classical methods, and can be readily integrated into existing MBES standard workflow. To facilitate future research, the code and pretrained model are available online
\footnote{The code and pretrained model are made available at \textcolor{red}{\href{https://github.com/luxiya01/mbes-score-denoise}{https://github.com/luxiya01/mbes-score-denoise}}.}.
\end{abstract}

\begin{IEEEkeywords}
multibeam echo-sounder, bathymetry mapping, point cloud denoising, point cloud outlier detection
\end{IEEEkeywords}

\begin{figure*}[h!]
    \centering
    \includegraphics[width=\textwidth]{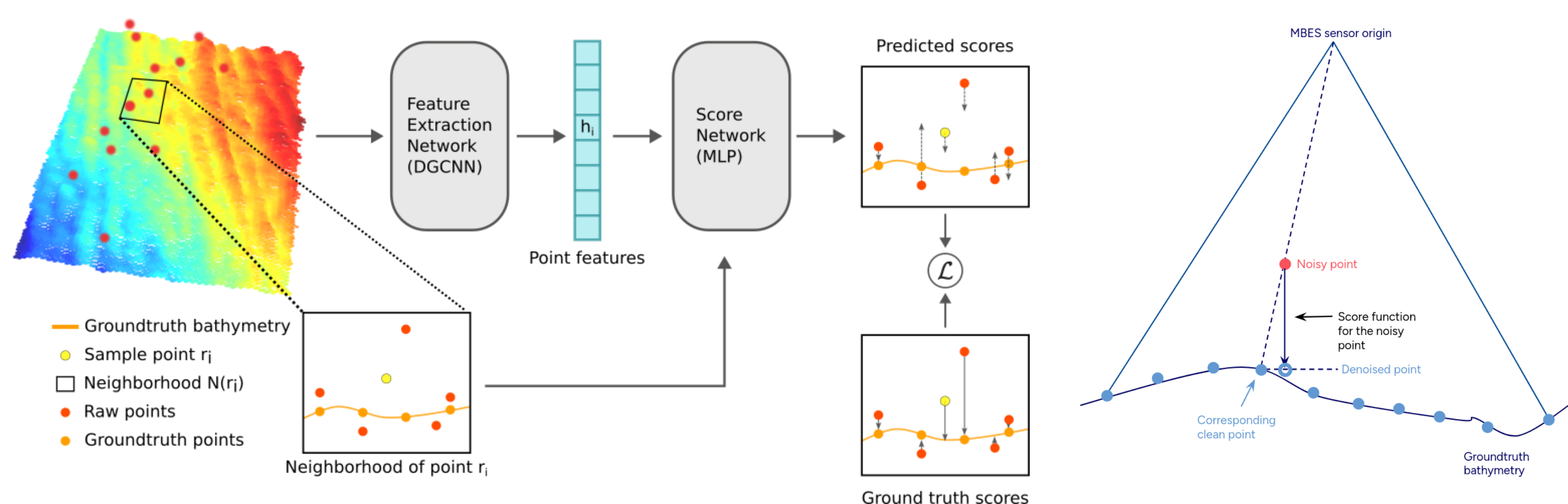}
    \caption{Left: Illustration of the score estimation network applied to an MBES patch. Right: An intuitive description of the \textit{score} function used to denoise MBES points. In this case, the score essentially represents the 1D vector from the \textit{z} value of the noisy point to the \textit{z} value the corresponding clean point.}
    \label{fig:scorenet}
\end{figure*}

\section{Introduction}

Detailed bathymetric information is the foundation of understanding many environmental processes in the ocean and at ocean boundaries \cite{morlighemBedMachineV3Complete2017}. In modern days, such information is primarily collected by multibeam echo-sounder (MBES) -- a sensor that can be mounted on surface vessels, remotely operated vehicles (ROVs) or autonomous underwater vehicles (AUVs). Thanks to technological improvements that result in cheaper and smaller MBES sensors, as well as global initiatives such as Seabed 2030 GEBCO project \cite{mayerNipponFoundationGEBCO2018}, the volume of available MBES data has grown exponentially \cite{ledeunfReviewDataCleaning2020}. However, raw MBES data contain between 1 - 25\% of outliers \cite{ledeunfReviewDataCleaning2020}.
%These outliers need to be detected in order to construct an accurate bathymetric model.
%Existing cleaning algorithms, such as Combined Uncertainty and Bathymetric Estimator (CUBE) \cite{calderAutomaticProcessingHighrate2003}, often require extensive manual parameter adjustment, with limited scalability and repeatability \cite{longComprehensiveDeepLearningBased2023}. Recently, two deep learning based models for MBES outlier detection have been proposed \cite{stephensUsingThreeDimensional2020, longComprehensiveDeepLearningBased2023}. However, neither of the two models compare their method to existing outlier detection algorithms. With the lack of open-sourced code and pretrained weights, it is difficult to fairly evaluate the performance of these models.
To handle these outliers, the standard bathymetric construction workflow typically combines semi-automatic cleaning algorithms, such as Combined Uncertainty and Bathymetric Estimator (CUBE) \cite{calderAutomaticProcessingHighrate2003}, with extensive parameter tuning and validation by data processing experts. Such workflow lacks scalability and repeatability.

% \begin{figure}[h!]
%     \centering
%     \includegraphics[width=.3\textwidth]{images/schematics.png}
%     \caption{An intuitive description of the \textit{score} function used to denoise MBES points. In this case, the score essentially represents the 1D vector from the \textit{z} value of the noisy point to the \textit{z} value the corresponding clean point.}
%     \label{fig:schematics}
% \end{figure}

In this work, we draw inspirations from the point cloud denoising community, and propose a score-based multibeam denoising network based on \cite{luoScoreBasedPointCloud2021}. In this case, the \textit{score} refers to the gradient of the log-probability function of the points. To train the score network, we used real AUV survey data and followed the standard workflow of manual data cleaning and mesh construction to create a ground truth dataset. This dataset is then used to evaluate our model's performance on MBES outlier detection and denoising, where \textit{denoising} refers to moving the noisy points closer to the clean surface. Compared to simple radius and statistical baselines, we found that this network is capable of achieving higher precision and recall on the outlier rejection task. For denoising, we noticed that the scores alone were not capable of fully recovering the true position of extreme outliers. We then developed an extension with classical mean interpolation technique to further improve the denoising results.
%To handle this situation, we extended the network with mean interpolation and further iterations of score-based denoising were required to achieve desirable results.

The main contributions of this work are as follows:
\begin{enumerate}
    \item We provide an open-source implementation of score-based MBES point cloud denoising method, and compare its performance to classical outlier detection and denoising methods.
    \item We propose a procedure that follows the standard MBES data pipeline to create ground truth dataset. As such, the proposed score-based training can be integrated into existing standard workflow.
\end{enumerate}

\section{Related Work} \label{sec:related-work}
\subsection{Outlier Detection on MBES Point Clouds}
Many semi-automatic methods have been developed for outlier detection on MBES point clouds. However, a survey paper from 2020 \cite{ledeunfReviewDataCleaning2020} concludes that performance comparison between these methods are difficult, as each are developed for difference scenarios and tested on different datasets. CUBE \cite{calderAutomaticProcessingHighrate2003} is arguably the most widely used semi-automatic algorithm for MBES cleaning and has been integrated into many commercial software. This estimator provides multiple depth hypotheses and the associated uncertainties on a pre-defined coordinate grid. Outliers are treated as new hypotheses, and manual inspection is required to reject the outliers. Due to this hypothesis tracking framework, CUBE performs poorly in chaotic seafloor \cite{ledeunfReviewDataCleaning2020}. Further, the initialization of CUBE uncertainties requires detailed knowledge of the survey system, including properties such as patch test, sensor offsets and attitude accuracy \cite{hare1995accuracy}, which are not always obtainable.

Recently, two deep learning methods have been proposed for MBES outlier rejection. In \cite{stephensUsingThreeDimensional2020}, MBES point clouds are downsampled and voxelized into 3D volumes, and a 3D UNet \cite{ronneberger2015unet} is trained to detect outliers. Due to the 3D voxelization, this method is computationally inefficient. Further, it is unable to distinguish between individual soundings and cannot handle too rapidly changing terrains \cite{stephensUsingThreeDimensional2020}. In \cite{longComprehensiveDeepLearningBased2023}, a simplified PointCleanNet \cite{rakotosaonaPointCleanNetLearningDenoise2019} is used to predict outlier probability directly from MBES point clouds. Clean data is augmented with manual noise and segmented into overlapping patches with 500 data points each. For each patch, the network predicts the outlier probability for the central 16 points, which tends to be either 0 or 1. The final outlier probability of a point is then the mean from several patch predictions. Notably, neither of the two deep learning methods compares their performance to existing outlier detection algorithms. With the lack of open-sourced code and pretrained weights, it is difficult to fairly evaluate the performance of these models.

\subsection{Point Cloud Denoising}
Denoising is the task of moving noisy points closer to the underlying clean surface. Although uncommon in MBES processing, this process is crucial for many 3D rendering applications. Traditionally, denoising is often formulated as an optimization problem, where to goal is to satisfy various geometric constrains, such as local density \cite{zaman2017density} and surface smoothness\cite{alexa2001point}. Deep learning based denoising can be divided into three categories: displacement-based method \cite{rakotosaonaPointCleanNetLearningDenoise2019, hermosilla2019total}, downsample-upsample methods \cite{luo2020differentiable} and score-based methods \cite{luoScoreBasedPointCloud2021}. Displacement-based methods predict the point-wise displacement in one step, and can lead to over-smoothing (shrinkage) and under-smoothing (outliers). Downsample-upsample methods suffer from shrinkage due to the loss of details during downsampling. In comparison, score-based methods iteratively denoise the point cloud using gradient ascent, alleviating the severity of both shrinkage and outliers \cite{luoScoreBasedPointCloud2021}.

\section{Method}
\subsection{MBES Sensor Basics}
MBES emits fan-shaped sound waves beneath the transducer and maps a wide swath of bathymetry from a single transmitted ping through beamforming of received signals. 
%Due to the beam steering process, the received beam lies on a cone underneath the sensor, with main axis aligned to the vehicle's forward direction, and the peak of the cone under the sensor's frame of origin.
Through a series of motion compensation, sound velocity adjustments and bottom detection, MBES outputs a set of 3D points representing the location where each beam detects the seafloor intersection.

In this work, we assume each beam forms a plane with the seafloor. This means that the X values, or the along-track values of the 3D points, are determined solely by the sensor and vehicle pose, whilst the Y and Z values are together determined by the range of bottom detection. Note that this assumption is a simplification. Due to beam steering, the principle response axis of the received beams is modified and the received beam pattern is warped into a conical shape \cite{lurton2002introduction}.

%For the MBES point cloud, the beams are focused to the surface of an infinite cone, with the main axis typically aligned to the vehicle's forward direction, and the peak of the cone located at the sensor's frame origin. In the simplest case, this cone can be reduced to a disk (i.e points lie in a  plane). In such case, the X values, or the along-track values, are determined by the sensor and vehicle pose, whilst the Y and Z values are together determined by the range of the bottom detection. In this work, we consider this simplest case. We assume that the X value of the points are fixed, and define a 1D score for the Z dimension. Note that the Y values can be moved according to MBES sensor geometry given Z.

\subsection{Problem Formulation}
Given a \textit{raw} MBES point cloud $\mathbf{R} = \{\mathbf{r}_i\}_{i=1}^N$, where each $\mathbf{r}_i \in \mathbb{R}^3$ consists of the XYZ hits of the MBES beam onto the seafloor, the goal is to recover the corresponding \textit{clean} MBES point cloud $\mathbf{C} = \{\mathbf{c}_i\}_{i=1}^N$. In this work, we assume that the X (along-track) values of $\mathbf{R}$ are fixed, and learn to denoise the point cloud by moving the Z (elevation) values. Note that the Y (across-track) values can be moved according to MBES sensor geometry given Z.

To denoise the point cloud, we adapt the score-based denoising from \cite{luoScoreBasedPointCloud2021}. More specifically, we assume that the clean set $\mathbf{C}$ is sampled from an underlying distribution $p$, whilst the noisy set $\mathbf{R}$ is sampled from $p$ convolved with a noise distribution $n$, $p*n$, plus additional outliers $o$. When the outliers are removed, the mode of $p*n$ represents the noise-free surface. The gradient of the log-probability of $p*n$, also known as the \textit{score} of $p*n$, can be denoted by $\nabla_\mathbf{r}\log[(p*n)(\mathbf{r})]$ and can be learned directly from $\mathbf{R}$. Using the \textit{score}, we can denoise $\mathbf{R}$ by performing gradient ascent, moving the noisy points to the mode of $p*n$. Moreover, the relative magnitude of the score can be used for outlier detection.

\subsection{Score Estimation Network}
\autoref{fig:scorenet} visualizes the components of the score estimation network.
For a noisy point $\mathbf{r}\in\mathbb{R}^3$, we define the ground truth score in the $z$ component as:
\begin{equation}
s(\mathbf{r}_z) = NN_{xy}(\mathbf{r}, \mathbf{C})_z - \mathbf{r}_z    
\end{equation}
where $NN_{xy}(\mathbf{r}, \mathbf{C})$ denotes the nearest neighbor to point $\mathbf{r}$ in $\mathbf{C}$ when only XY values are considered, and $[]_z$ denotes the $z$ component of the point. Intuitively, this score represents the 1D vector from the noisy $z$ component to its noise-free correspondence (see the \textit{Right} subfigure in \autoref{fig:scorenet}).

A score estimation network inspired by \cite{luoScoreBasedPointCloud2021} is then used to learn the \textit{local} score function $S_i(\mathbf{r})$, which corresponds to the gradient field around a point $\mathbf{r}_i$. For any point $\mathbf{r} \in \mathbb{R}^3$ in the vicinity of $\mathbf{r}_i$, its score relative to $\mathbf{r}_i$ is computed as follows:
\begin{equation}
    \mathcal{S}_i(\mathbf{r}) = \text{MLP}(\mathbf{r}-\mathbf{r}_i, \mathbf{h}_i)
\end{equation}
where MLP denotes multi-layer perceptron, $\mathbf{r} - \mathbf{r}_i$ denotes the relative position of $\mathbf{r}$, and $\mathbf{h}_i$ denotes the feature of $\mathbf{r}_i$ extracted by a DGCNN \cite{wangDynamicGraphCnn2019}, a feature extraction network that considers both local and global context.

To train the network, we optimize the L2-norm between the estimated score $S_i(\mathbf{r})$ and ground truth score $s(\mathbf{r}_z)$:
\begin{equation}
    \mathcal{L}^{(i)} = \mathbb{E}_{\mathbf{r}\sim\mathcal{N}(\mathbf{r}_i)}[||s(\mathbf{r}_z)-S_i(\mathbf{r})||_2^2]
\end{equation}
with $\mathcal{N}(\mathbf{r}_i)$ representing the local neighborhood of point $\mathbf{r}_i$. The final training objective is then the average of all localized losses of the inlier set $\mathcal{I}$:
\begin{equation}
    \mathcal{L} = \frac{1}{|\mathcal{I}|}\sum_{i\in \mathcal{I}} \mathcal{L}^{(i)}
\end{equation}
%Note that outliers are not used for score learning, since their local score signals contradict that of the inliers.
Note that only the local score function of the inliers are used for loss computation.

\subsection{Score-Based Denoising for MBES} \label{sec:score-denoising}
After training, the local score network $S_i(\mathbf{r})$ can be used to estimate the score of any point. 
%For a noisy raw point $\mathbf{r}$, the final predicted score is the ensemble score of its neighbors:
%For any noisy point $\mathbf{r}$, the final predicted score is the median of all scores for point $\mathbf{r}$ computed using the local score functions of its neighbors:
For a raw point $\mathbf{r}$, we first compute the score of $\mathbf{r}$ using its $k$ nearest neighbors' local score functions. Then, the median of these predicted scores is used as the final score $\mathcal{E}(\mathbf{r})$:
\begin{equation} \label{eq:ensemble}
    \mathcal{E}(\mathbf{r}) = \text{median}_{\mathbf{r}_i\in kNN(\mathbf{r})} S_i(\mathbf{r})
\end{equation}
Instead of the original \textit{mean} function used in \cite{luoScoreBasedPointCloud2021}, we choose \textit{median} as the ensemble function to mitigate the influence of clusters of severe outliers in MBES data.

\subsubsection{Outlier Detection}
Given the ensemble score of all raw points, we use \textit{interquartile range} from descriptive statistics \cite{dekking2006statistics} to detect outliers. Specifically, we compute the 25 percentile $Q_1$ and 75 percentile $Q_3$ scores per MBES patch. The interquartile range $IQR$ is given by $Q_3 - Q_1$, and points with score below $Q_1 - i*IQR$ and above $Q_3 + i*IQR$ are treated as outliers. In this work, $i$ is set to 5 for best outlier detection results.

\subsubsection{Denoising}
Using the ensemble score, the $z$ component of the noisy point $\mathbf{r}_i$, denoted as $z_i$, can be iteratively denoised for $T$ steps as follows: let  $z_i^{(0)} = z_i$, the subsequent $z_i^{(t)}$ are computed using $z_i^{(t)} = z_i^{(t-1)} +\alpha_i\mathcal{E}(z_i^{(t-1)})$, where $\alpha_i$ is the step size at step $t$, and $\mathcal{E}$ is an ensemble score function given by \autoref{eq:ensemble}. We follow \cite{luoScoreBasedPointCloud2021} and decay $\alpha$ exponentially as the denoising step $t$ increases, i.e. $\alpha_t = \alpha_0 \times \gamma^t$. In this work, we used the default setting from \cite{luoScoreBasedPointCloud2021}, and set $\alpha_0$ to $0.2$, the common ratio $\gamma$ to $0.95$ and $t$ to 30 steps.

\section{Experiments}
\subsection{Datasets} \label{sec:dataset}
\begin{figure}[h!]
    \centering
    \includegraphics[width=.6\linewidth]{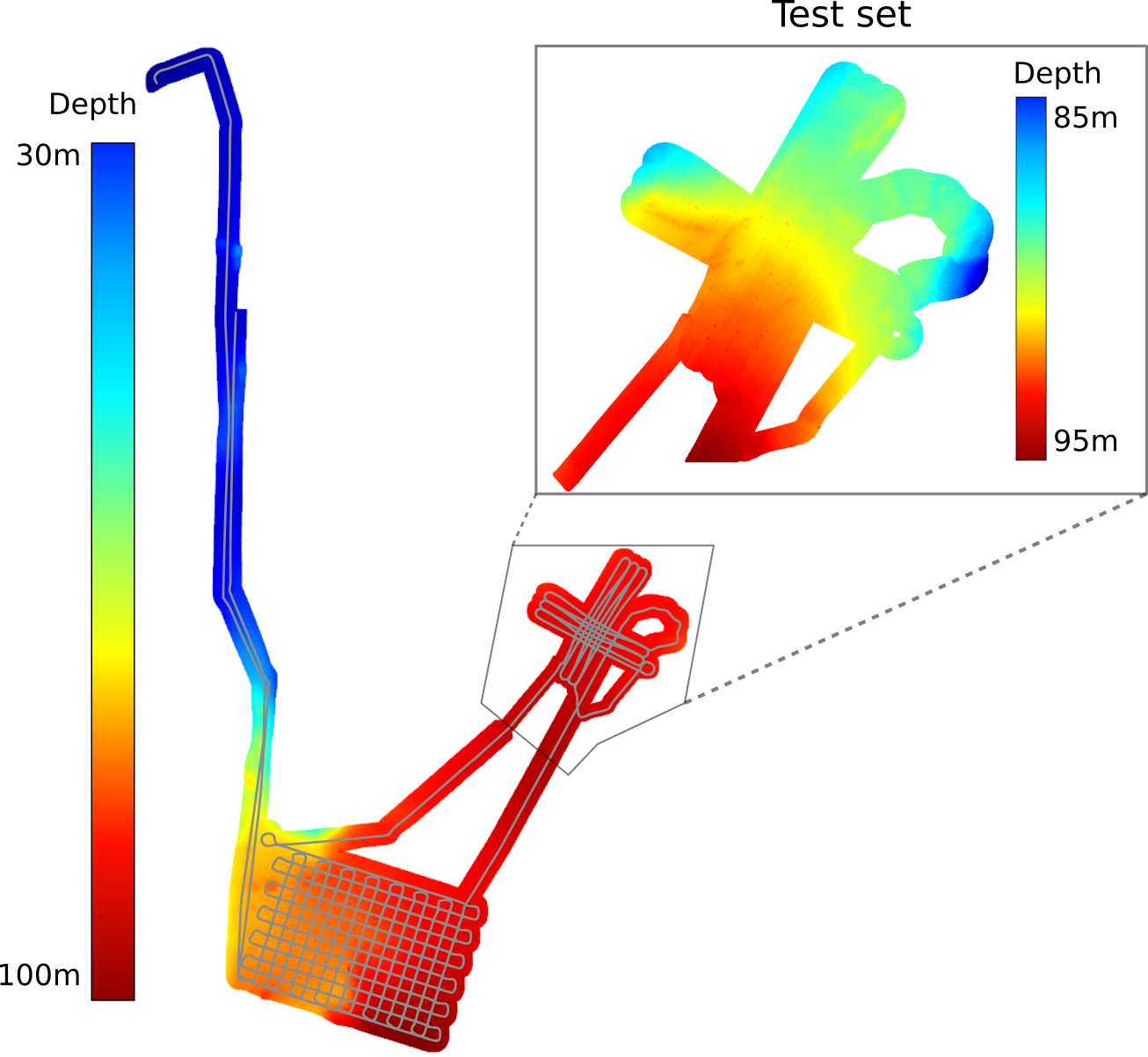}
    \caption{Visualization of MBES dataset used in this paper. The AUV trajectory is overlaid on the cleaned bathymetry in gray. The test set is enlarged and highlighted on the top right corner. Note that the depth range in the test set (85-95m) is significantly different than that of the training set (30-100m).}
    \label{fig:dataset}
\end{figure}

\begin{figure}[h!]
    \centering
    \includegraphics[width=\linewidth]{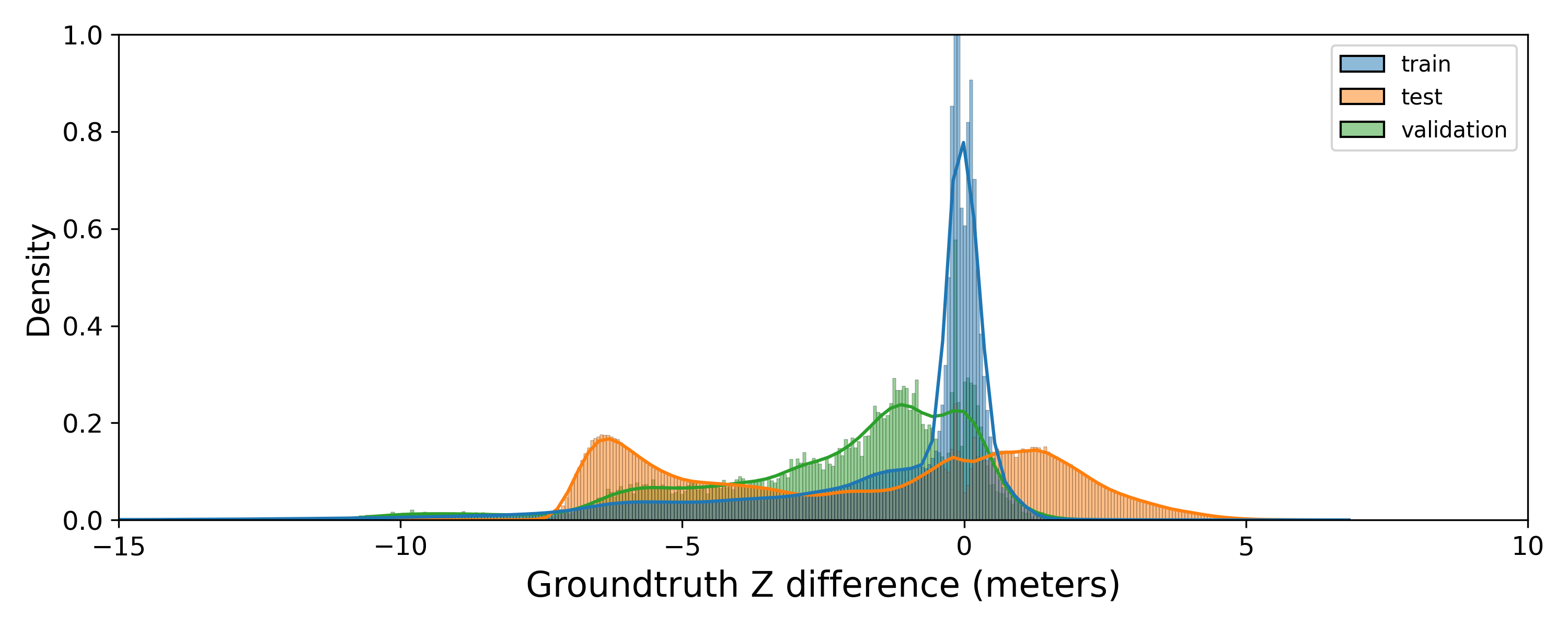}
    \caption{Distribution of $z$ value differences between the manually identified outliers in the raw MBES data and clean draping results in training, test and validation set.\vspace{-20pt}}
    \label{fig:raw_z_diff}
\end{figure}

A big obstacle in developing and evaluating MBES denoising methods is the lack of ground truth. In this work, we manually clean a MBES dataset collected by a Hugin AUV with Kongsberg's EM2040 sensor in an area rich of trawling patterns. This dataset covers around 5h of survey time and contains 216k MBES pings, each with 400 beams. The cleaned bathymetry and the vehicle trajectory are visualized in \autoref{fig:dataset}. Further details of the dataset are given in \autoref{tab:dataset}. After cleaning, we create a mesh using \href{https://www.eiva.com/products/navisuite/navisuite-processing-software/navimodel-producer}{EIVA NaviModel}, and use \href{https://github.com/nilsbore/auvlib}{AUVLib}'s draping functionality to construct the ground truth noise-free point cloud. To construct the denoising dataset, we then divide the data into 32-ping patches, resulting in 6777 patches for training and evaluation. Finally, this dataset is divided into two geographically separate sections for training and testing. A small section of the training set is used for validation. From the number of manually detected outliers in the entire dataset and the test set (see \autoref{tab:dataset}), as well as the distribution of raw $z$ differences shown in \autoref{fig:raw_z_diff}, it is evident that the outlier characteristics of the test set is significantly different from that of the training set.

Before feeding into the score network, each MBES patch is demeaned and then normalized. Since the XY values represent the geographical UTM coordinates and the Z values represent the bathymetry, these values have different ranges. As such, we separate the normalization of XY and Z. For Z, we compute the global maximum Z values in the training and test set separately, and normalize the demeaned Z values by dividing over the set maximum. For XY, we normalize each patch individually, so that all points fall into a unit sphere. Finally, to increase the diversity of the data, we randomly sample rotations $\in [0-180\degree]$ around $z$-axis during training, simulating random change in yaw angles.

\begin{table}[h!]
\centering
\caption{Details of the denoising dataset.}
\label{tab:dataset}
  \begin{tabular}{@{}ll@{}}
    \toprule
    Details & Specifications \\
    \midrule
    Vehicle speed & 2 m/s \\
    Vehicle altitude & $\sim$ 20 m \\
    Survey duration & $\sim$ 4 h \\
    Sonar frequency & 400 kHz \\
    Ping rate & 2.5 Hz ($\sim$ 0.4 s/ping) \\
    Beam forming & 400 beams across 120\degree \\
    Total number of points & 86,710,800 points (217k pings) \\
    Total number of outliers & 3,410,764 points (3.93\% of all points)\\
    Number of test points & 25,518,400 points (64k pings) \\
    Number of test outliers & 1,880,800 points (7.37\% of test points)\\
    \bottomrule
  \end{tabular}
\end{table}

\subsection{Evaluation Metrics} \label{sec:eval}
We evaluate the proposed method for both outlier detection and denoising. For outlier rejection, we compute the common binary classification metrics, including precision, recall, accuracy and F1-score \cite{stephensUsingThreeDimensional2020, longComprehensiveDeepLearningBased2023}. For denoising, we report the point-wise Chamfer distance (CD), mean absolute error (MAE) and root-mean-squared error (RMSE) in \textit{z} \cite{irisawaHighResolutionBathymetryDeepLearning2024}. All denoising metrics are computed after denormalizing the patches back to metric scale.

\subsection{Baselines} \label{sec:baseline}
For outlier rejection, we choose two baselines implemented in the \href{https://www.open3d.org/docs/latest/tutorial/geometry/pointcloud_outlier_removal.html}{Open3D library}:
\begin{itemize}
    \item \textit{Statistical outlier removal} computes the average distance of points to its neighbors in a point cloud, and treats points too far away from its neighbors as outliers.
    \item \textit{Radius outlier removal} counts the number of neighboring points within a sphere of given radius. It treats points with few neighbors as outliers.
\end{itemize}
To ensure strong baseline performance, we perform hyperparameter search on the entire test set, and use F1-score to select the best performing parameters for both baselines. \textit{Statistical outlier removal} achieves the highest F1-score when computing average distance using 30 neighbors and detecting outliers at $std$ of 1.5. \textit{Radius outlier removal} achieves highest F1-score with a radius of 0.03 on a normalized patch, outliers are detected if less than 30 neighbors are contained within the sphere.

For denoising, we first identify outliers using the best parameter settings of the baseline outlier rejection methods. The Z values of the identified outliers are then estimated using two interpolation methods: 1) the mean Z of the 16 neighboring points; 2) Ordinary Kriging \cite{chilesGeostatisticsModelingSpatial2012} implemented in the  \href{https://geostat-framework.readthedocs.io/projects/pykrige/en/stable/}{PyKrige package} with default parameters.

%For outlier rejection, we evaluate Ordinary Kriging \cite{chilesGeostatisticsModelingSpatial2012} used by \cite{irisawaHighResolutionBathymetryDeepLearning2024}, PointCleanNet \cite{rakotosaonaPointCleanNetLearningDenoise2019} used by \cite{longComprehensiveDeepLearningBased2023}, as well as classical point cloud filtering methods such as statistical outlier removal and radius outlier removal.

\subsection{Network Training}
For the score network, we train for 10500 steps with a batch size of 4 MBES patches. Adam optimizer was used with an initial learning rate of 1e-4. The learning rate is halved when the improvement in MAE$_z$ stagnates on the validation set. The trained network is evaluated under three settings, with 64, 128 and 256 neighbors used for ensemble score computation (see \autoref{eq:ensemble}), respectively.

\section{Results}
\subsection{Outlier Rejection}
\begin{figure*}
    \centering
    \includegraphics[width=\linewidth]{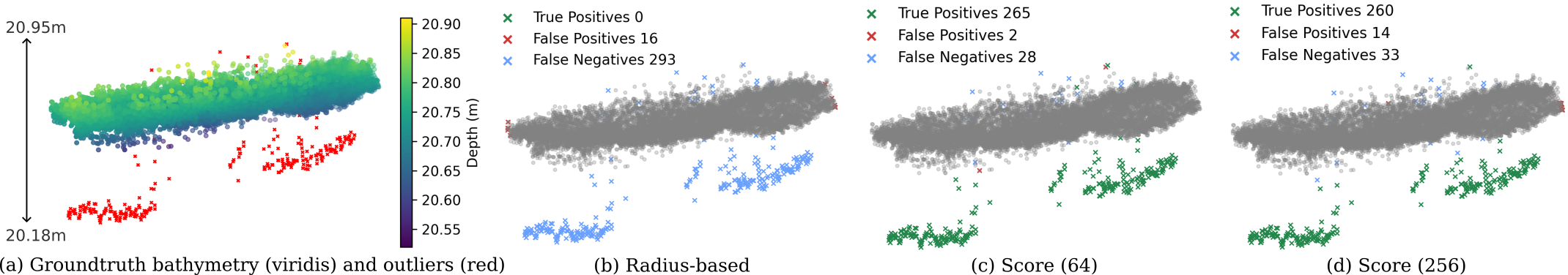}
    \caption{Visual outlier detection results of a patch where the proposed score net achieves better results than baseline methods (radius-based method is visualized here). Typically, these patches have very small range in $z$ compared to the rest of the test set. As a result, the baselines with globally tuned parameters performs poorly. The score-based methods, on the other hand, learn the local score around points and are less sensitive to the exact numeric range within a patch.\vspace{-5pt}}
    \label{fig:score-outlier-success}
\end{figure*}

\begin{figure*}
    \centering
    \includegraphics[width=.8\linewidth]{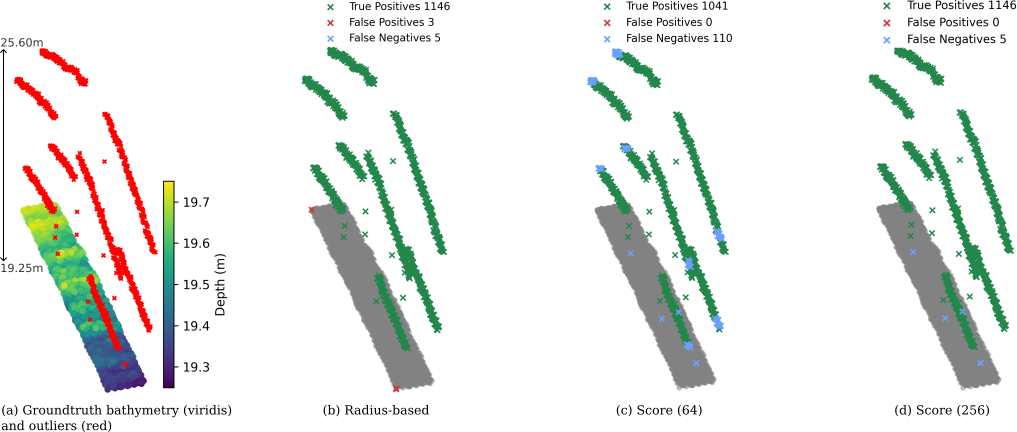}
    \caption{Visual outlier detection results of a patch where Score (64) performs worse than the baselines (radius-based method is visualized here). Score (256) tends to outperform Score (64). Typically, these patches have large range of $z$, and the outliers are significantly removed from the inliers. In the patch shown, the outliers range between 19.25-25.6m, whilst inliers reside within 19.3-19.7m. Due to the large amount of outliers closed to each other, the estimated scores will be inflated, leading to worse performance of score-based methods. This inflation can partly be mitigated by using larger neighborhood for score ensemble (see differences between Score (64) in (c) and Score (256) in (d)).}
    \label{fig:score-outlier-failure}
\end{figure*}

\autoref{tab:outlier-results} shows the outlier rejection results. Since the test set contains $\sim 7.37\%$ outliers (see \autoref{tab:dataset}), a naive classifier that treats all points as inliers will achieve an accuracy of $92.63\%$ with $0\%$ precision and recall. From \autoref{tab:outlier-results}, it is evident that the score network can successfully be trained for MBES outlier detection. Compared to the two baseline methods, all three settings of the score networks achieve higher accuracy, precision, recall and F1-score. Within the three settings, Score (256) -- the setting with highest number of neighbors for ensemble score computation, achieves overall best results. This is because many of the outliers in the test set are clustered together. As such, a larger neighborhood for ensembling leads to more robust outlier detection. \autoref{fig:score-outlier-success} and \autoref{fig:score-outlier-failure} showcase a typical success and failure case for score-based outlier detection in our dataset, respectively.

\begin{table}[h!]
    \centering
    \caption{Outlier rejection results. The best and second best method per metric are highlighted in \textcolor{red}{red} and \textcolor{magenta}{magenta}, respectively.}
    \begin{tabular}{lllll}
    \toprule
         Method & Accuracy & Precision & Recall & F1-score \\ 
         \midrule
         Radius & 98.49\% & 98.79\% & 80.61\% & 0.8878 \\
         Statistical & 98.17\% & 90.76\% & 83.70\% & 0.8709 \\
         \midrule
         Score (64) & 99.22\% & \textcolor{magenta}{99.67\%} & 89.78\% & 0.9447 \\
         Score (128) & \textcolor{magenta}{99.46\%} & \textcolor{red}{99.72\%} & \textcolor{magenta}{92.91\%} & \textcolor{magenta}{0.9620} \\
         Score (256) & \textcolor{red}{99.50\%} & 99.31\% & \textcolor{red}{93.91\%} & \textcolor{red}{0.9653} \\
    \bottomrule
    \end{tabular}
    \label{tab:outlier-results}
\end{table}

\subsection{Denoising}
\begin{figure}
    \centering
    \includegraphics[width=\linewidth]{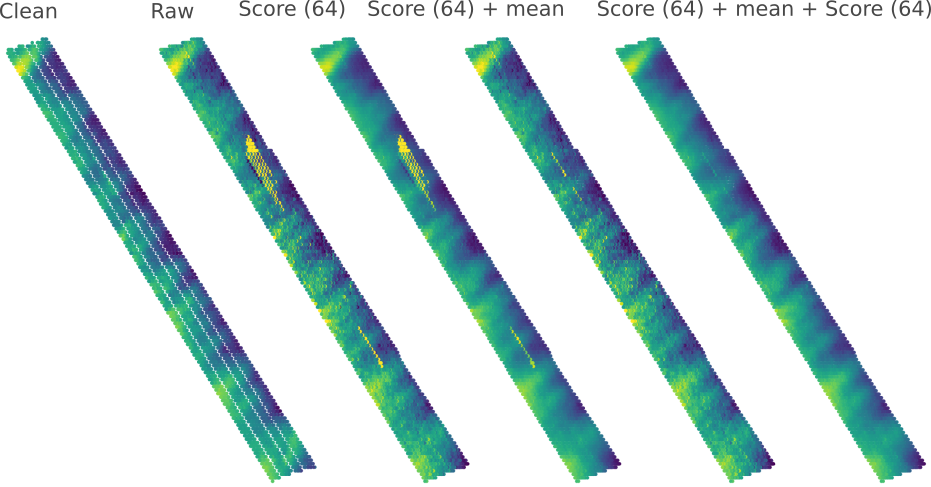}
    \caption{Example results using Score (64) with three different denoising procedures.\vspace{-0pt}}
    \label{fig:score-denoising}
\end{figure}

The denoising results on denormalized patches are shown in \autoref{tab:denoising-results}. For the score networks, we experimented with three denoising procedures, each with 64, 128 and 256 neighbors for score ensemble:
\begin{itemize}
    \item Score (knn): Denoising is performed solely using the learned scores (see \autoref{sec:score-denoising}).
    \item Score (knn) + mean: Outliers are detected using Score (knn). The $z$ values of the outliers are interpolated the mean of 16 closest neighbors.
    \item Score (knn) + mean + Score (knn): Outliers are detected using Score (knn) and interpolated with neighbor means. This modified patch without outliers is then fed into Score (knn) again for final denoising. 
\end{itemize}
An example using the above three denoising procedures is visualized in \autoref{fig:score-denoising}.

\begin{figure*}
    \centering
    \includegraphics[width=.9\linewidth]{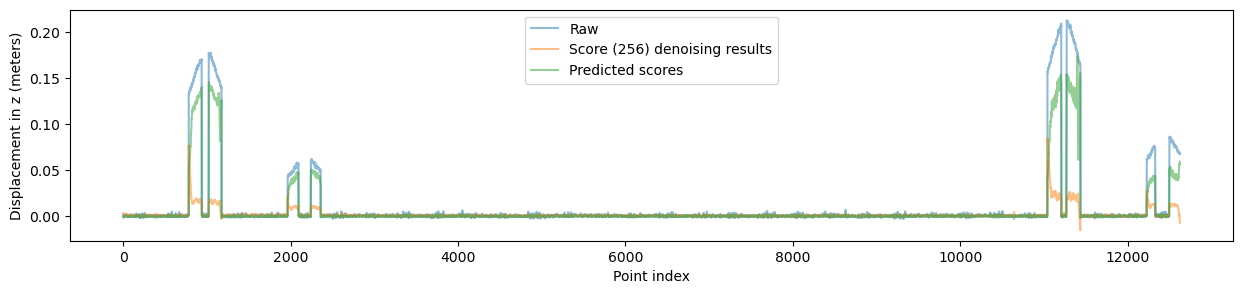}
    \caption{Example MBES patch denoising results using Score (256).\vspace{-10pt}}
    \label{fig:score-gradients}
\end{figure*}

From \autoref{tab:denoising-results}, we notice that although the pure score-based denoising [Score (knn)] are capable of partially reducing all three metrics, adding mean interpolation and another score-denoising iteration further improves the results. This indicates that the magnitudes of the predicted scores are not large enough to fully cover the extent of extreme outliers, which are common in the test set (see outlier distributions in \autoref{fig:raw_z_diff}). An example of this phenomenon can be seen in the score visualization in \autoref{fig:score-gradients}. In this figure, although the predicted scores align well with the original displacement in $z$, the outliers are still clearly visible after denoising with Score (256).

When comparing the mean interpolation in combination with outlier rejection, both baselines achieve lower denoising metrics than any score-based methods (Score (knn) + mean), despite the score-based methods outperforming both baselines in all outlier rejection metrics (see \autoref{tab:outlier-results}). This difference is more pronounced in CD and RMSE$_z$, two metrics that use L2-norm and thus square the errors. This is probably due to the score-based methods sometimes failing to identify extreme outliers (See \autoref{fig:score-outlier-failure}).

\begin{table}[h!]
    \centering
    \caption{Denoising results. The best and second best methods per metric are highlighted in \textcolor{red}{red} and \textcolor{magenta}{magenta}, respectively.}
    \begin{tabular}{llll}
    \toprule
    Method & CD & MAE$_z$ & RMSE$_z$ \\ \midrule
    Raw & 1.1058 & 0.2418 & 0.9136 \\
    \midrule
    Radius + mean & \textcolor{magenta}{0.07983} & 0.02788 & \textcolor{magenta}{0.06153} \\
    Statistical + mean & 0.08127 & 0.02719 & 0.06242 \\
    \midrule
    Radius + Ordinary Kriging & 0.08173 & 0.02838 & 0.06292 \\
    Statistical + Ordinary Kriging & 0.08173 & 0.02774 & 0.06394 \\
    \midrule
    Score (64) & 0.2773 & 0.06483 & 0.3748 \\
    Score (128) & 0.2037 & 0.05478 & 0.2806 \\
    Score (256) & 0.1201 & 0.04405 & 0.1714 \\
    \midrule

    Score (64) + mean & 0.2450 & 0.05471 & 0.3368 \\
    Score (128) + mean & 0.1640 & 0.03994 & 0.2301 \\
    Score (256) + mean & 0.08348 & 0.02397 & \textcolor{red}{0.05759} \\
    \midrule

    Score (64) + mean + Score (64) &  0.08790 & 0.02327 & 0.1043 \\
    Score (128) + mean + Score (128) & 0.08416 & \textcolor{magenta}{0.02208} & 0.09158 \\
    Score (256) + mean + Score (256) & \textcolor{red}{0.07907} & \textcolor{red}{0.02049} & 0.07416 \\
    \bottomrule
    \end{tabular}
    \label{tab:denoising-results}
\end{table}

%\newpage
\section{Conclusions}
In this paper, we adapted a score-based point cloud denoising model for outlier detection and denoising of real MBES survey data, and proposed a pipeline that can be readily integrated into existing MBES data processing workflow. For outlier detection, all score-based models outperform both the radius and statistical baselines. For denoising, we noticed that despite being able to reduce the noise, extreme outliers are still visible (though with smaller displacements) after direct application of score-based denoising.
To tackle this shortcoming, we developed an extension that incorporates classical mean interpolation and another iteration of score denoising. The final model exhibits better or on-par denoising performance compared to the classical baselines.
%However, the results can be improved by incorporating mean interpolation and another iteration of score denoising.
Further studies are needed to validate the generalizability of the model.

\section*{Acknowledgement}
This work was supported by Stiftelsen för Strategisk Forskning (SSF) through the Swedish Maritime Robotics Centre (SMaRC) (IRC15-0046). The data collection was performed using ship Askholmen and the Ran AUV, owned and operated by the University of Gothenburg. The AUV was funded by Knut and Alice Wallenberg foundation through MUST (Mobile Underwater System Tools). The computations were enabled by the supercomputing resource Berzelius provided by National Supercomputer Centre at Linköping University and the Knut and Alice Wallenberg Foundation, Sweden.

\bibliographystyle{IEEEtranBST/IEEEtran}
% \bibliography{auv_ref}
\bibliography{auv_ref_nolink}

\end{document}